\documentclass{bmvc2k}

\newcommand{\ie}{{\it i.e.}}
\newcommand{\eg}{{\it e.g.}}
\newcommand{\etal}{{\it et al.}}
\newcommand{\Tref}[1]{Table~\ref{#1}}

\newcommand{\Fref}[1]{Fig.~\ref{#1}}

\newcommand{\Cref}[1]{Chapter~\ref{#1}}
\newcommand{\omni}{$360^\circ$}
\usepackage{multirow}
\usepackage{arydshln}
\usepackage{textcomp}

\makeatletter

\makeatletter

\title{Intersection Prediction from Single \omni~Image via Deep Detection of Possible Direction of Travel}

\addauthor{Naoki Sugimoto}{naoki48916@gmail.com}{1}
\addauthor{Satoshi Ikehata}{sikehata@nii.ac.jp}{2}
\addauthor{Kiyoharu Aizawa}{aizawa@hal.t.u-tokyo.ac.jp}{1}

\addinstitution{
 The University of Tokyo\\
 Tokyo, Japan
}
\addinstitution{
 National Institute of Informatics\\
 Tokyo, Japan
}

\runninghead{Student, Prof, Collaborator}{BMVC Author Guidelines}

\def\eg{\emph{e.g}\bmvaOneDot}

\def\etal{\emph{et al}\bmvaOneDot}

\begin{document}

\maketitle

\begin{abstract}
Movie-Map, an interactive first-person-view map that engages the user in a simulated walking experience, comprises short~\omni~video segments separated by traffic intersections that are seamlessly connected according to the viewer's direction of travel. However, in wide urban-scale areas with numerous intersecting roads, manual intersection segmentation requires significant human effort. Therefore, automatic identification of intersections from \omni~videos is an important problem for scaling up Movie-Map.
In this paper, we propose a novel method that identifies an intersection from individual frames in \omni~videos. Instead of formulating the intersection identification as a standard binary classification task with a~\omni~image as input, we identify an intersection based on the number of the possible directions of travel (PDoT) in perspective images projected in eight directions from a single~\omni image detected by the neural network for handling various types of intersections. 
We constructed a large-scale \omni~Image Intersection Identification (iii360) dataset for training and evaluation where \omni~videos were collected from various areas such as school campus, downtown, suburb, and china town and demonstrate that our PDoT-based method achieves 88\% accuracy, which is significantly better than that achieved by the direct 
%n\"aive 
binary classification based method. The source codes and a partial dataset will be shared in the community after the paper is published.
\end{abstract}

%-------------------------------------------------------------------------

\section{Introduction}

\begin{figure}[t]
 \begin{center}
  \begin{minipage}{0.24\hsize}
        \begin{center}
          \includegraphics[keepaspectratio,width=\textwidth]{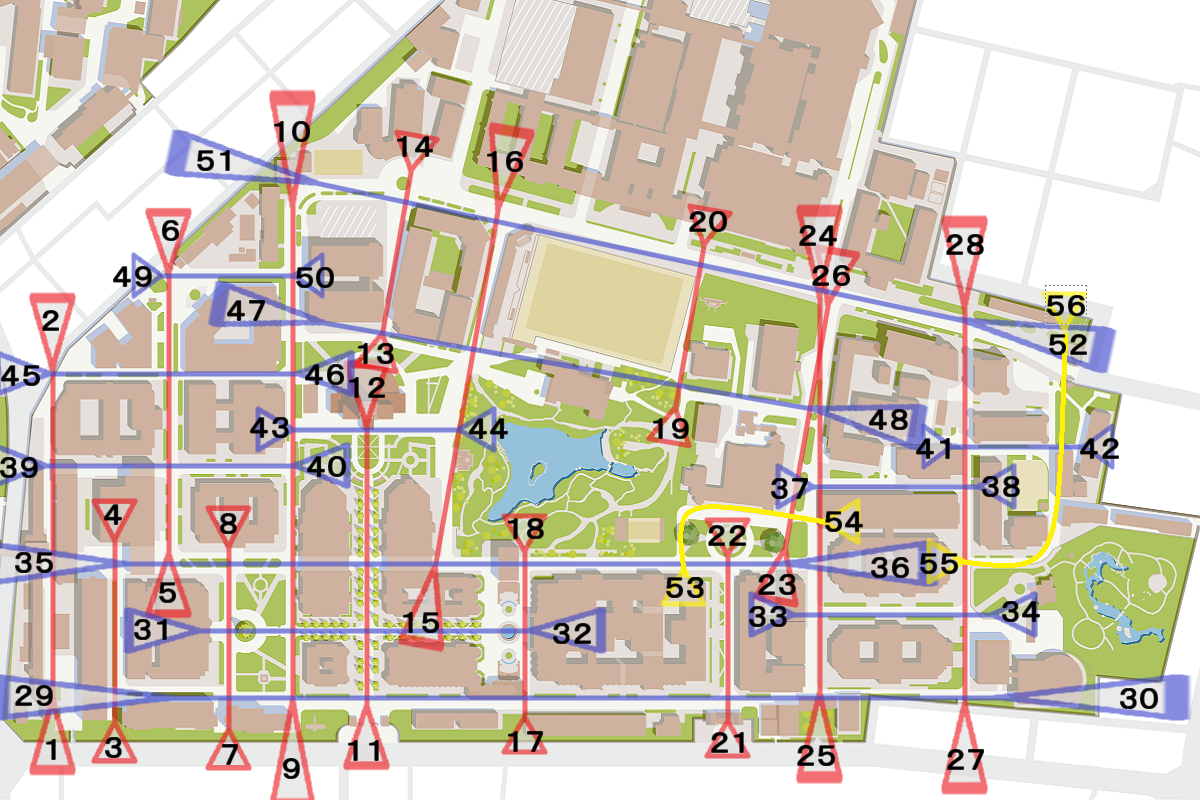}
          \vspace{2pt}
          (a) Campus
        \end{center}
    \end{minipage}
  \begin{minipage}{0.24\hsize}
        \begin{center}
          \includegraphics[keepaspectratio,width=\textwidth]{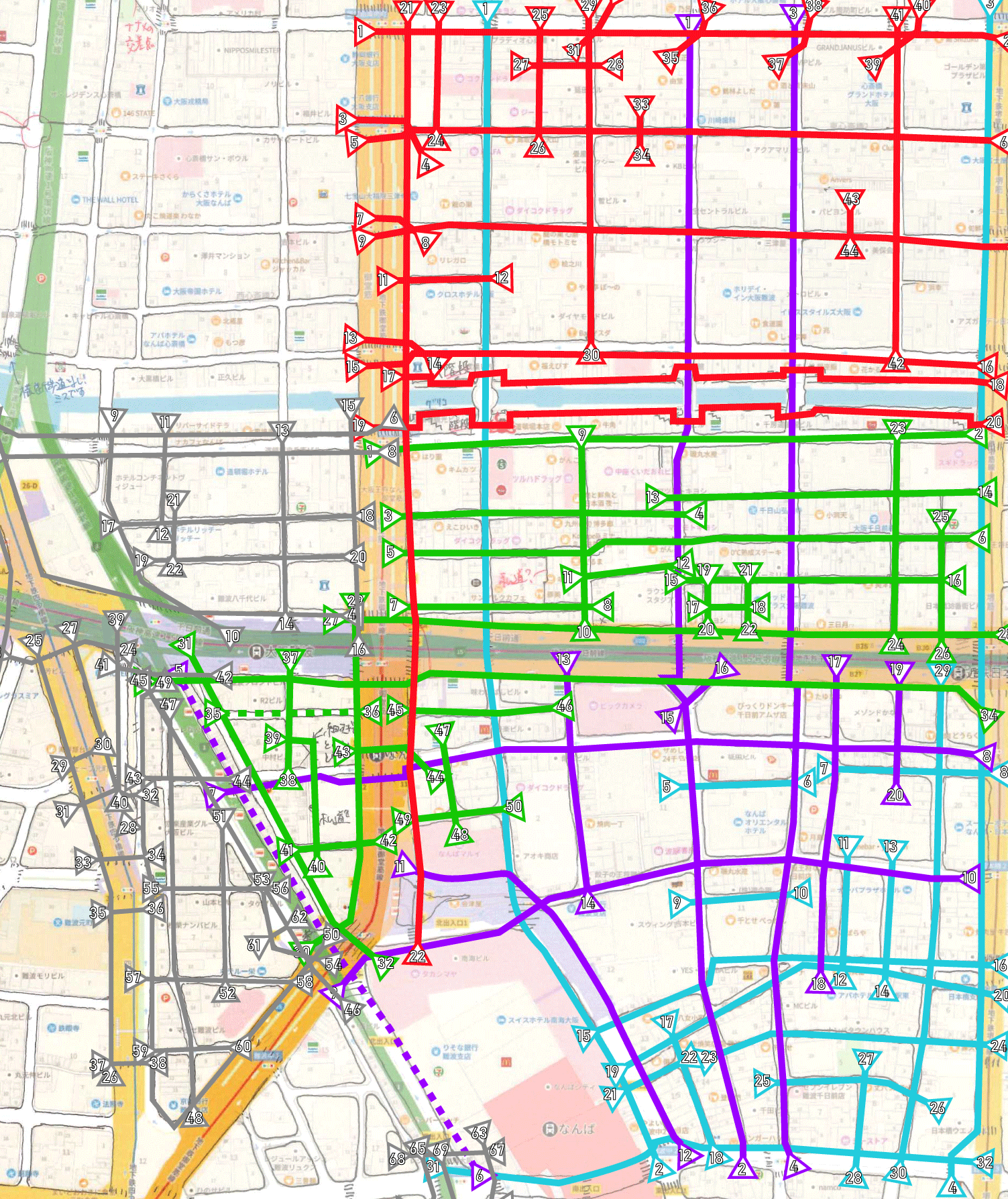}
          
          \vspace{2pt}
          (b) Downtown
        \end{center}
    \end{minipage}
  \begin{minipage}{0.24\hsize}
        \begin{center}
          \includegraphics[keepaspectratio,width=\textwidth]{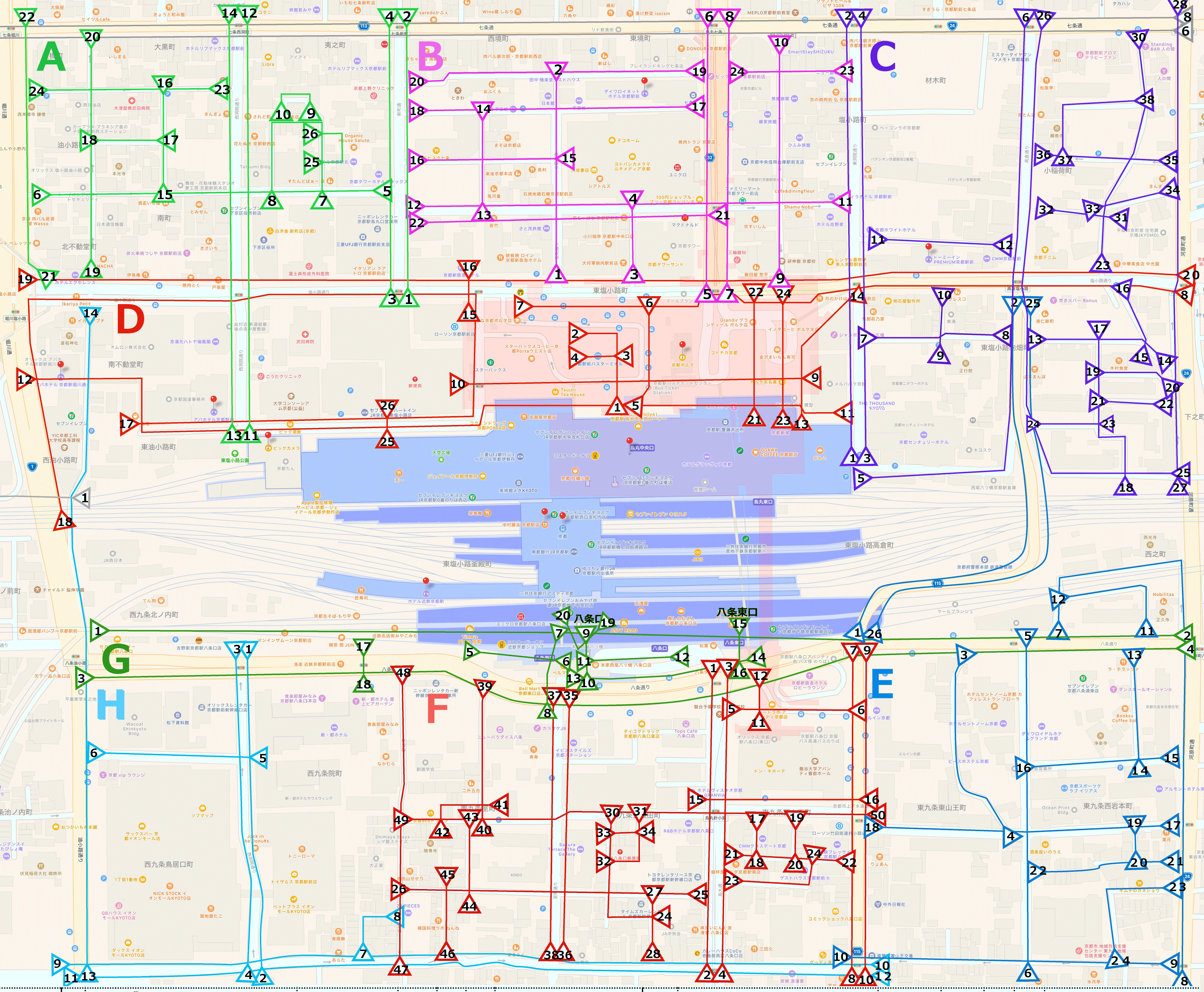}
          
          \vspace{2pt}
          (c) Suburb
        \end{center}
    \end{minipage}
  \begin{minipage}{0.24\hsize}
        \begin{center}
          \includegraphics[keepaspectratio,width=\textwidth]{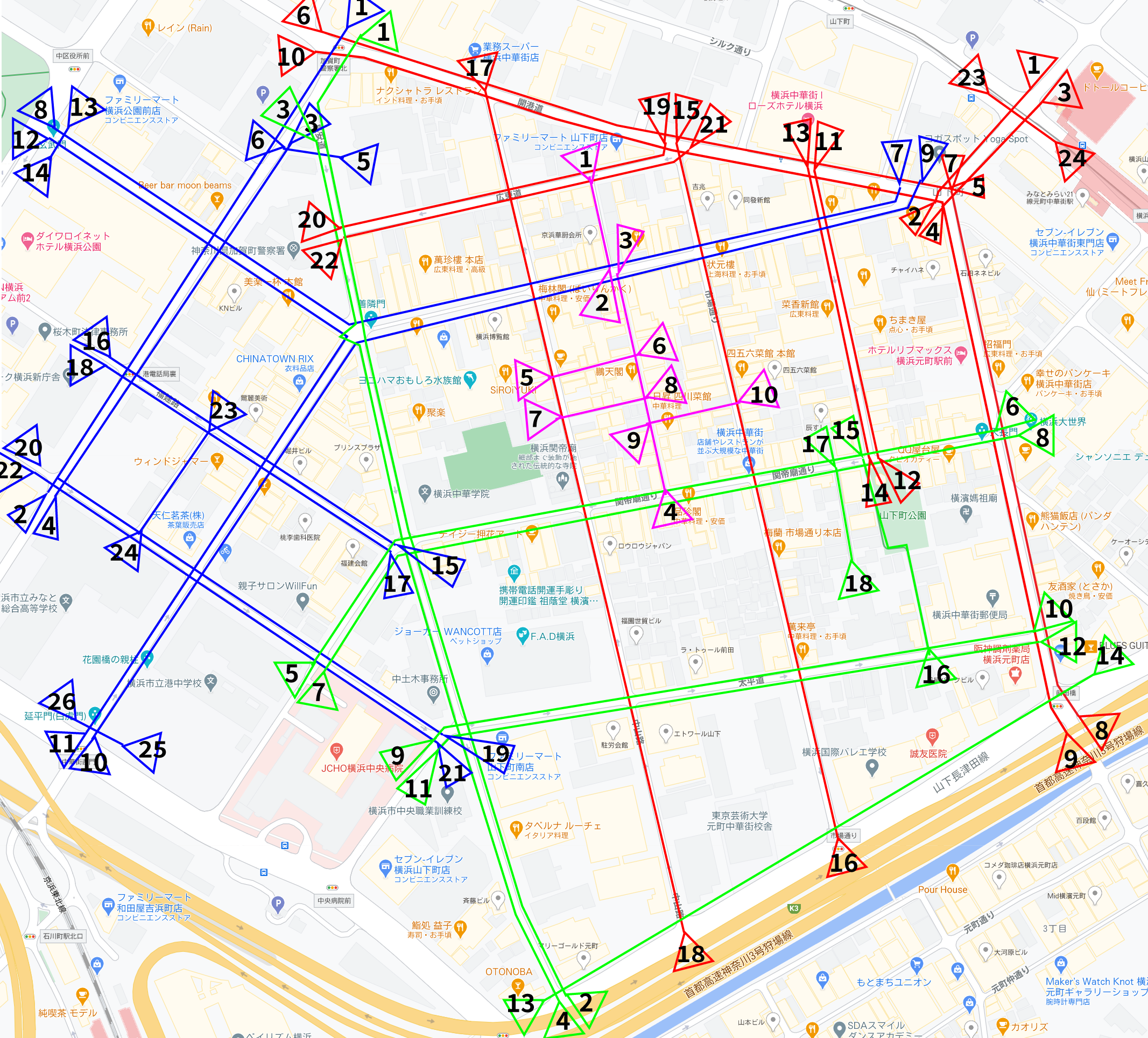}
          
          \vspace{2pt}
          (d) Chinatown
        \end{center}
    \end{minipage}
  \caption{Routes and intersections in the collected \omni walk-around videos. The arrows and triangles on each map indicate the video paths and intersections, respectively.}
  \label{moviemap_routes}
  \end{center}
\end{figure}

%Since ancient times, maps have been useful tools as a method of expressing geographical information~\cite{mapfan,offmaps,geographical_survey}. The recent digitalization of maps has made it possible to hold a lot of information together on a map. For instance, it has become possible to present first-person images of a specific location on a map at the same time, giving the viewer the immersive experience of actually being there~\cite{google_map,mapillary,map_advantage}. However, it is often difficult to get the same information as when you actually walk around the place by just looking first-person static images or bird's-eye view. In such a case, {\it Movie-Map}~\cite{movie_map,sugimoto_acmmm20} is an attractive tool, which is an interactive map with first-person view to engage the user in a simulated walking through an unfamiliar scenes. Interaction in Movie-Map is very simple: a first-person \omni~walk-around video is presented in real time according to the user's position on the map. The user can interactively change the walking direction and speed as well as the view direction.
{\it Movie-Map}~\cite{movie_map,sugimoto_acmmm20} is a digital map application that presents first-person images of a specific location on a map, giving the viewer the immersive experience of actually being there~\cite{google_map,mapillary,map_advantage}. Internally, Movie-Map comprises short \omni~video segments separated by traffic intersections. When passing through an intersection, the video segments are seamlessly connected according to the direction of travel, and the user remains unaware of the connection. However, in reality, shooting a large amount of short videos between intersections is inefficient; therefore street-level videos are typically captured and then split by intersections. In the existing works, this intersection segmentation has been performed manually~\cite{movie_map} or by deciding the intersections after SLAM of each street video~\cite{sugimoto_acmmm20}. However, as shown in~\Fref{moviemap_routes}, because of the presence of several intersections in the areas targeted by Movie-Map, manual intersection segmentation requires significant effort. Furthermore, dynamic objects such as pedestrians or cars and less-textured landscapes frequently hamper the accuracy of image-based SLAM~\cite{slam1}. In addition, a comparison between camera poses and visual information from intersecting video sequences of complex streets is not always accurate.  
%other which restricts the diversity of cities, towns, and villages that Movie-Map can handle. 
Although GPS can be used to obtain the coordinates of the camera locations, the error is sufficiently large to prevent correct localization. Moreover, GPS is not applicable to indoor situations, either.

In this paper, we propose a learning-based algorithm to identify the intersection automatically using a single \omni~video frame. The intersection frame was used to split walk-around videos into short video segments. The most challenging condition in this task is the use of a single image; if accomplished, it can be used most generally; that is, we can determine the intersections without using information from other video sequence frames.
%We make intersection decisions on a single image, not a sequence of images because it is possible to increase the amount of data and learning by collecting and annotating \omni~images intersection images, even if there are no images of the route taken as shown in our experiment. 
%Rather than directly identifying intersection images as the standard classification task, 
We formulated this problem as the detection of possible directions of travel (PDoT) for handling various types of intersections. Provided a \omni~frame sampled from a video, perspective projections for different views were applied to generate multiple perspective images (\ie, $8$ views in our experiments). Then, we classified the possibility of an observer to walk in the forward direction of the field-of-view (FoV) without being blocked by obstacles such as buildings. If PDoTs were observed in three or more views, the frame was identified as a traffic intersection. 
As a baseline for comparison, we also prepared a direct binary classification of the provided image without using PDoTs. 
The experimental results showed that our PDoT-based method can achieve 88\% accuracy in identifying a single image intersection and generally outperformed the direct classification-based approach for various real scenes. 

We trained our PDoT detection network and evaluated the intersection identification method on our new dataset, \omni Image Intersection Identification (iii360), which was generated from \omni~walk-around videos in four areas and \omni~images from the Google Street View (GSV) panoramas~\cite{street_view}. The change in accuracy of intersection identification was investigated with different combinations of training and testing datasets. The results demonstrated that the proposed method achieved 88\% accuracy even for an area not included in any of the three datasets used for training, whereas the n\"aive binary classification network achieved 65\% detection accuracy.

%While video segments extraction for Movie-Map is the most important %application of the results obtained that we have assumed, the intersection identification from a \omni~view could also be useful in the field of automated driving or robot navigation where an autonomous agent requires %to correctly recognize the traffic of people and automobiles and prepare %for unpredictable behavior near intersections.

\section{Related Work}

\subsection{Movie-Map}

{\it Movie-Map}~\cite{movie_map} was proposed in 1980 as the first interactive map to engage a user in a simulated driving experience. The original Movie-Map system was built 
%on a legacy computer, 
using an optical videodisc and four stop-frame film cameras -- the cameras, mounted on the top of a car, were triggered approximately every 10 ft. Movie-Map simulates travel by displaying controlled rate sequences of individual frames captured at periodic intervals along a particular street in a town. Despite being an innovative concept, the system was impractical because of the large human efforts involved. For instance, to allow the route to deviate from straight paths down each street, separate sequences were captured to display all the possible turns at every intersection. In addition, the captured videos had to be split manually by intersection to connect the different driving videos through those turn sequences. Owing to the lack of scalability caused by the large human effort as well as that of computational resources and data capacity at that time, Movie-Map was relegated to less importance until recently, and a system based on static \omni~images (such as GSV~\cite{street_view}) became the mainstream approach to digital map navigation.

Recently, Sugimoto~\etal~ redesigned Movie-Map with modern imaging and information processing technologies and demonstrated its superiority to GSV in exploring unfamiliar scenes~\cite{sugimoto_acmmm19,sugimoto_icmr20}. In addition to replacing large-camera and expensive disk systems in~\cite{movie_map} with consumer \omni~cameras and personal computers, they proposed a method to automate the time-consuming intersection segmentation. Specifically, they applied Visual SLAM~\cite{slam1} to recover and track the 3-D camera trajectories of the walk-around videos, aligned the trajectories onto the map, identified the intersections using the aligned trajectories, and refined them using visual features. However, their method relied heavily on the results of Visual SLAM, which is not robust to dynamic objects such as cars or people and texture-less landscapes. On the other hand, as our method does not rely on either 3-D reconstruction or multiple frames from the entire frame sequence, it is less sensitive to dynamic objects and low-textured scenes. 

\subsection{Traffic intersection detection}
In addition to Movie-Map, intersection identification using a pedestrian viewpoint or vehicle-mounted cameras has recently been actively studied in automated driving and robot navigation fields. Owing to the highly unpredictable behavior of traffic intersections, correct recognition and safe behavior in their proximity is essential for an autonomous agent. 

Depending on the input information, intersection identification methods are broadly categorized into two types: those that use non-visual information such as LIDAR or GPS data and those that use images or videos. Without using visual information, Fathi and Krumm~\cite{Fathi2010} identified intersections in an urban-scale traffic network using GPS data from several vehicles. On the other hand, Zhue~\etal~\cite{Zhu2012} proposed a method for intersection detection using sparse 3-D point clouds obtained from in-vehicle 3-D LIDAR data. Unfortunately, GPS data are frequently inaccurate in urban areas with tall buildings, and 3-D LIDAR is expensive and basically less portable than consumer cameras, which are inappropriate for capturing walk-around videos for some applications such as Movie-Map.

With the development of deep learning technologies, image-based methods for intersection identification are becoming more attractive owing to their cheap installation cost and compatibility with in-vehicle or robot-mounted cameras. Bhatt~\etal~\cite{Bhatt2017} formulated this task as a binary classification problem to identify intersections in short-time on-board videos and developed a variant of long-term recurrent convolutional network as the classifier. Although the classifier has been shown to be effective for in-vehicle videos, its reliability in other domains such as walk-around pedestrian-view videos, is still unclear. In addition, as this model accepts multiple video frames as input, its robustness in dynamic scenes with several walking people, such as in urban shopping malls, is questionable. On the other hand, Astrid~\etal~\cite{Astrid2020} recently proposed a single-image intersection classification system using ResNet-based architecture trained on a pedestrian-view-level image dataset containing 345 and 498 intersection and non-intersection images, respectively. Although the system achieved a test accuracy of 80\%, the training and test data comprised pedestrian-view images of sidewalks with little or no pedestrian traffic, and the generalization of this method to disparate test scenes from the training dataset is still not confirmed.

Unlike all these studies, the input of our method is a single frame from \omni~walk-around videos recorded for Movie-Map. To the best of our knowledge, this is the first attempt to identify an intersection from a single \omni~image. In addition, our targets include not only in-vehicle images or places with no pedestrians but also various cities, towns and villages with active people and car traffic, including places with different building designs. In such cases, conventional approaches such as direct binary classification of images are difficult to implement. Therefore, we propose a new intersection identification method based on the detection of PDoTs.
% \subsection{Deep Neural Network Leaning Using  \omni~Image}

% When using a  \omni~image as input to a Deep Neural Network (DNN), especially when applying a Convolution Neural Network (CNN)~\cite{cnn} to a distorted image, there are three ways to treat it: input them to the model as same as perspective images\cite{360_deep_original}, transform the kernel of the CNN to fit the distortion of the \omni~image~\cite{360_deep_kernel}, and use perspective projection method like cubemap~\cite{cubemap} and apply the model to each projected perspective images~\cite{360_deep_projection}.
% In our proposed method, the projecting way using visual vanishing points is adopted.
% In this paper, we discuss the difference in accuracy of intersection determination between the way that directly learns the relationship between the original \omni~image and whether it is in intersection or not and the proposed method.
\section{Proposed Method}

The purpose of this study was the automatic identification of intersection frames in walk-around videos shot for Movie-Map and the division of the video into intersection-wise segments. To achieve this goal, we propose a novel single \omni~image intersection detection algorithm that is applicable to challenging pedestrian views in various types of scenes.

The task is to classify whether a selected the image was captured at an intersection. The most straightforward method would be a data-driven approach wherein a large number of intersection and non-intersection labeled images are prepared to train deep neural networks, as performed in~\cite{Astrid2020}.

\begin{figure}[t]
  \begin{center}
  \includegraphics[keepaspectratio,scale=0.13]{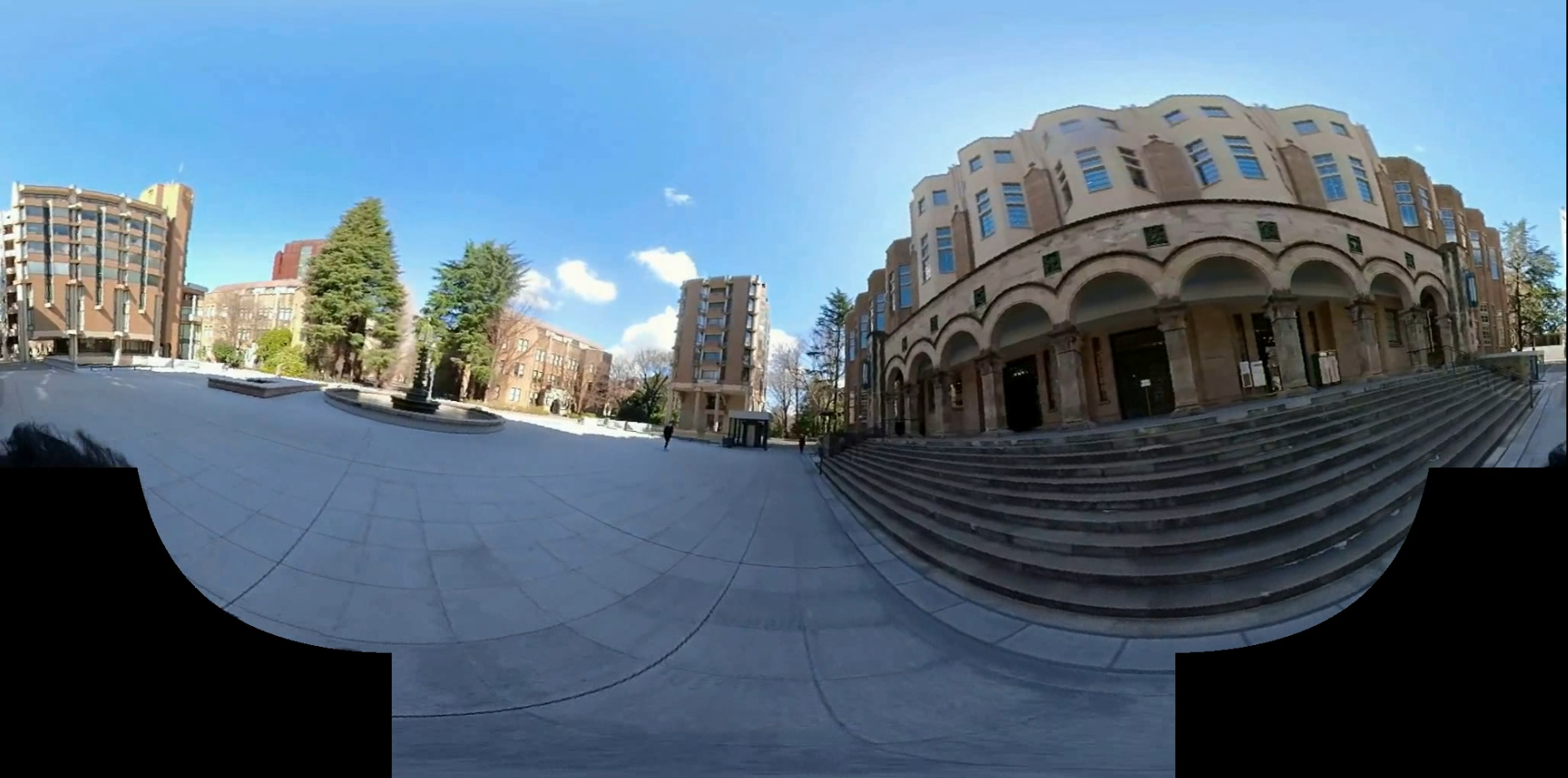}
  \caption{Example of an ambiguous intersection image. In this school campus scene, the road is omnidirectional, and the number of PDoTs is difficult to accurately identify.}
  \label{intersection_or_not}
  \end{center}
\end{figure}

Unfortunately, the definition of an intersection in reality is quite vague, except for a clearly sectioned roadway. For instance, a place with people moving in various directions, such as a university campus shown in~\Fref{intersection_or_not}, should be recognized as an intersection in practical applications such as Movie-Map because of the intersecting people and vehicles. However, even a human would hesitate to label this location as an intersection because of the absence of clearly sectioned road. Conventional intersection detection studies~\cite{Fathi2010,Zhu2012,Bhatt2017,Astrid2020} did not encounter this ambiguity as they only focused on on-board cameras or pedestrian view on clearly segmented sidewalks. 

\begin{figure}[t]
  \begin{center}
  \includegraphics[keepaspectratio,scale=0.2]{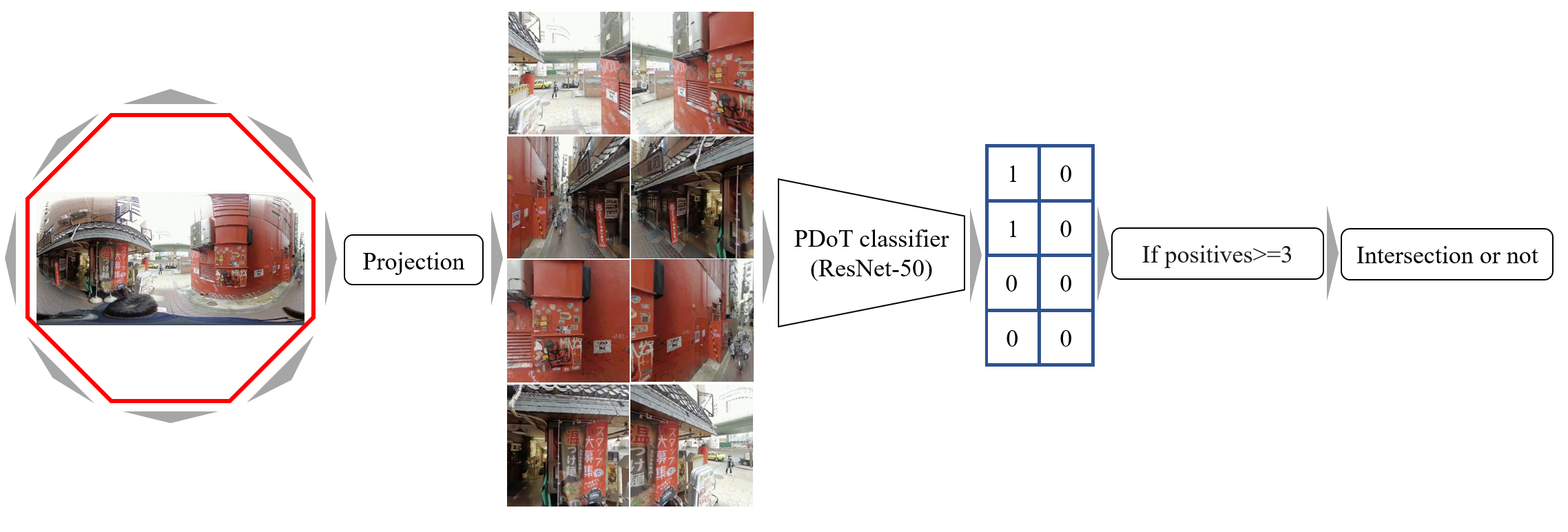}
  \caption{Illustration of the proposed method. Multiple non-overlapping perspective images of the $45^\circ\times 45^\circ$ FoV are cropped in eight view directions and our PDoT classifier is applied to individual images. If three or more PDoTs are detected, the \omni~image is classified as an intersection frame.}
  \label{proposed_method}
  \end{center}
\end{figure}

To address this problem, we redefined the intersection itself more rigorously and defined the problem accordingly. Specifically, an intersection in a \omni~image was defined as {\it a location where there are multiple PDoTs}. %(\ie, more than three in our experiments). 
Specifically, if a traveler can proceed in three and more directions, including the direction they have already been traveling, the location should be considered an intersection. The overview of our method is shown in~\Fref{proposed_method} If the direction of travel contains an obstacle such as a building, it is defined as not be travelable. Under this definition, both conventional (\eg, T,Y,X-intersections) as well as omnidirectional traffic intersections (\eg, those in a university campus square) are labeled as intersections. Notable, identification of omnidirectional intersections is crucial in several practical applications. 
Movie-Map needs to merge walk-around videos in two directions in a park square. 
%In Movie-Map, there are cases where we want to merge walk-around videos in two different directions in a park square. 
Advance detection of large spaces that are expected to contain heavy traffic is also important for robot navigation.

We propose a single \omni~image intersection classification network based on the redefined intersection. Deep neural networks for various tasks using \omni~images have been studied extensively in recent years and can be mainly divided into three approaches: apply neural networks directly to equirectangular projection (ERP) images~\cite{360_deep_original}, define kernels for convolution on a sphere~\cite{360_deep_kernel}, or divide a \omni~image into multiple perspective projection images and apply neural networks to them individually~\cite{360_deep_projection}. While the ERP-based method is attractive owing to its simplicity, our experiment validated that this direct approach is not always effective for a wide variety of real-world intersections (including omnidirectional intersections) because of large distortions in a \omni~image caused by sphere-to-plane projection. Therefore, we formulate this problem as the detection of multiple PDoTs in sampled perspective views, rather than directly identifying the intersections in \omni~images.

Specifically, we converted a single \omni~image provided in the ERP format (\eg, a single frame of a \omni~walk-around video) into multiple non-overlapping normal perspective images. 
The FoV of each perspective image should not be quite small or quite large because it is insufficient or contains multiple PDoTs, respectively.
%If the field-of-view of each perspective image is too small, the information contained in one projected image is insufficient, while the images with too large field-of-view may contain multiple PDoTs. 
To balance the trade-off between redundant and insufficient information, we cropped eight perspective images of the $45^\circ\times 45^\circ$ FoV from a target~\omni~image.
For each perspective image, we classified whether the center of the FoV contained a PDoT by applying a variant of the ResNet-50~\cite{resnet} network that will be explained in the implementation details. The input \omni~image was identified as an intersection if a minimum of three PDoTs were observed in the \omni~field-of-view. By dividing the complex problem of direct intersection identification from the entire \omni~view into a set of problems that determine the presence of PDoT in the narrow FoV, the system displays greater robustness to scene divergence, which was validated by the experimental results. Furthermore, by converting a \omni~image in the ERP format to perspective images, our method is theoretically less sensitive to sphere-to-plane projective distortions. Notable, although our method requires multiple model inference to determine whether a single \omni~ image is an intersection, they are independent and can be executed in parallel. Moreover, as perspective local images are much smaller than the original \omni~image, the computational cost is quite manageable.
 
%\begin{figure}[t]
%  \begin{center}

%  \includegraphics[keepaspectratio,scale=0.35]{figure/fvpp_detail2.png}
%  \caption{Sampling from a \omni~image in inference of FVPP.}
%  \label{fvpp_detail}
%  \end{center}
%\end{figure}

\section{iii360 Dataset}
The identification of an intersection in a single \omni~image is a new problem, with consequent nonavailability of training and test datasets. Therefore, we constructed a large-scale \omni~Image Intersection Identification (iii360) dataset for training our PDoT detection network and evaluating our method.

\subsection{Training data generation}

\begin{figure*}[t]
 \begin{center}
  \begin{minipage}{0.24\hsize}
        \begin{center}
          \includegraphics[keepaspectratio,width=\textwidth]{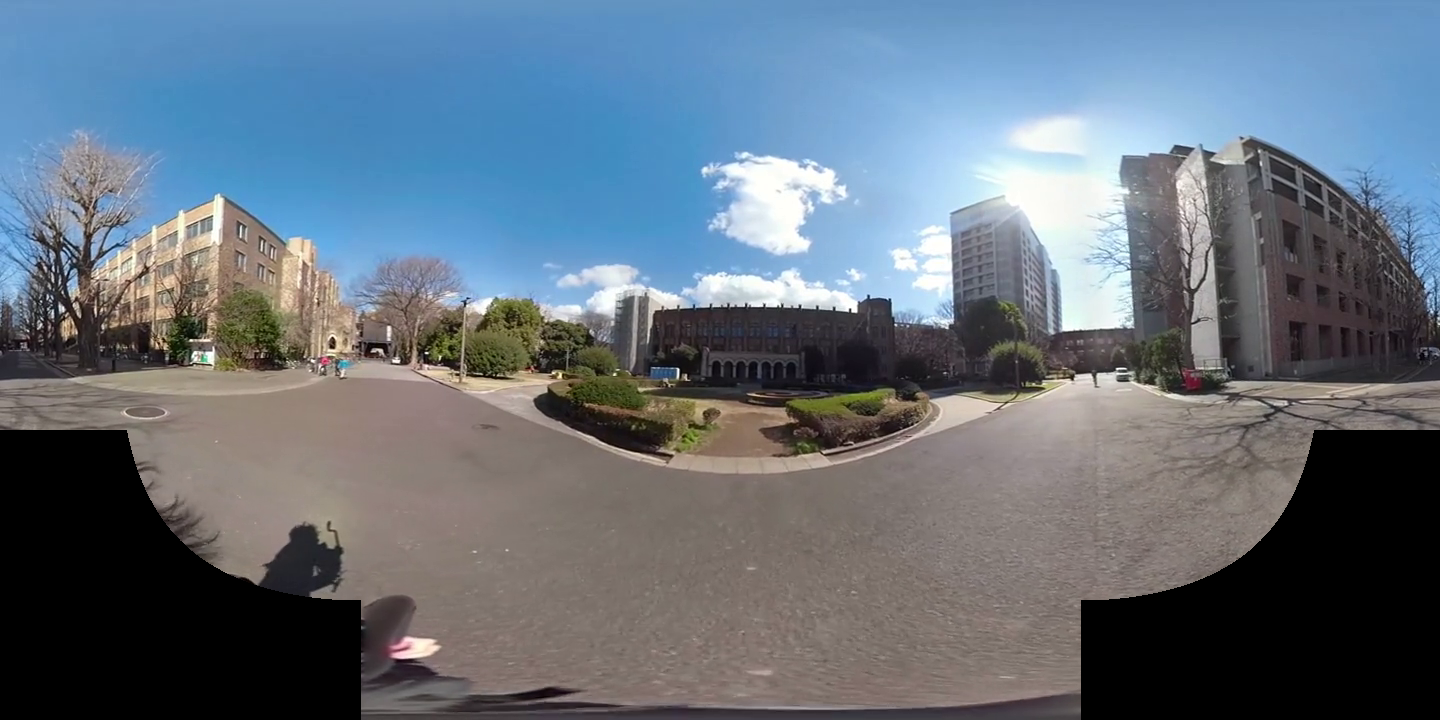}

          \vspace{2pt}
          (a) Campus
        \end{center}
    \end{minipage}
  \begin{minipage}{0.24\hsize}
        \begin{center}
          \includegraphics[keepaspectratio,width=\textwidth]{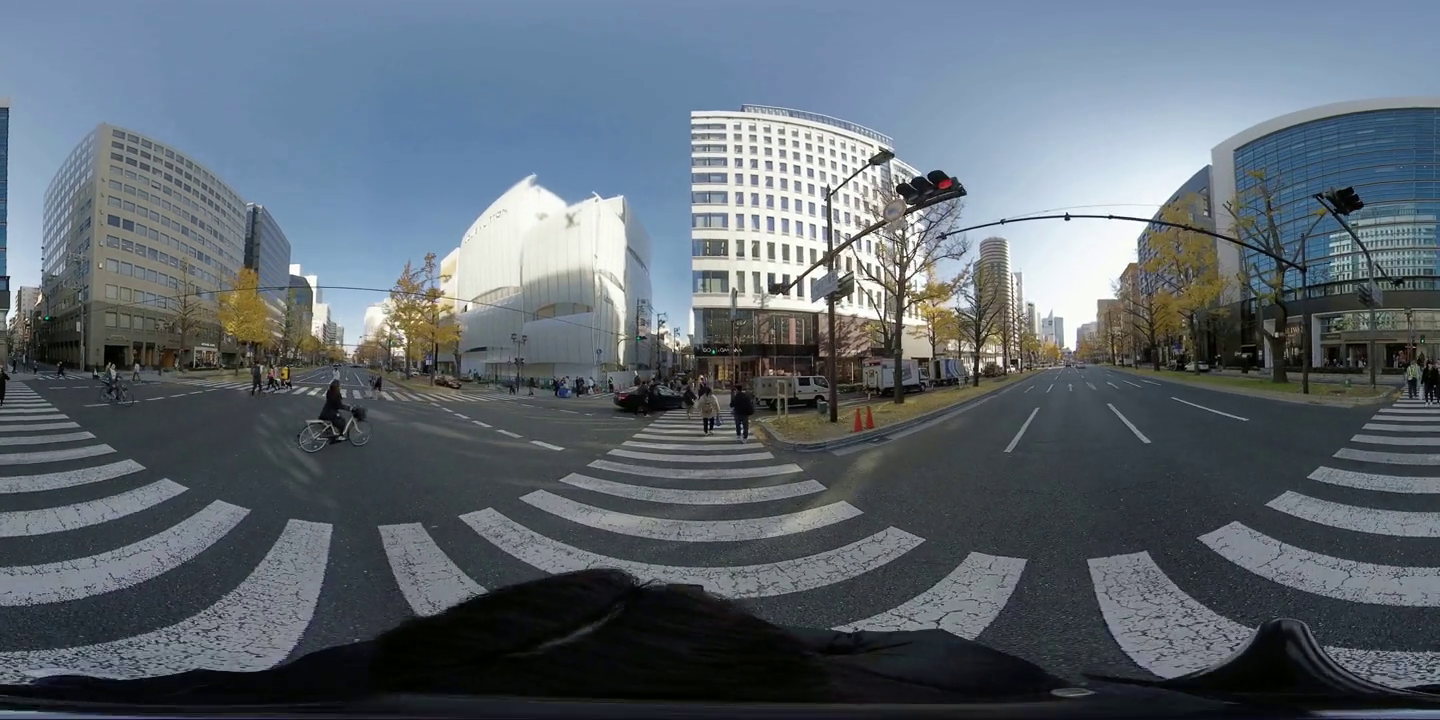}

          \vspace{2pt}
          (b) Downtown
        \end{center}
    \end{minipage}
  \begin{minipage}{0.24\hsize}
        \begin{center}
          \includegraphics[keepaspectratio,width=\textwidth]{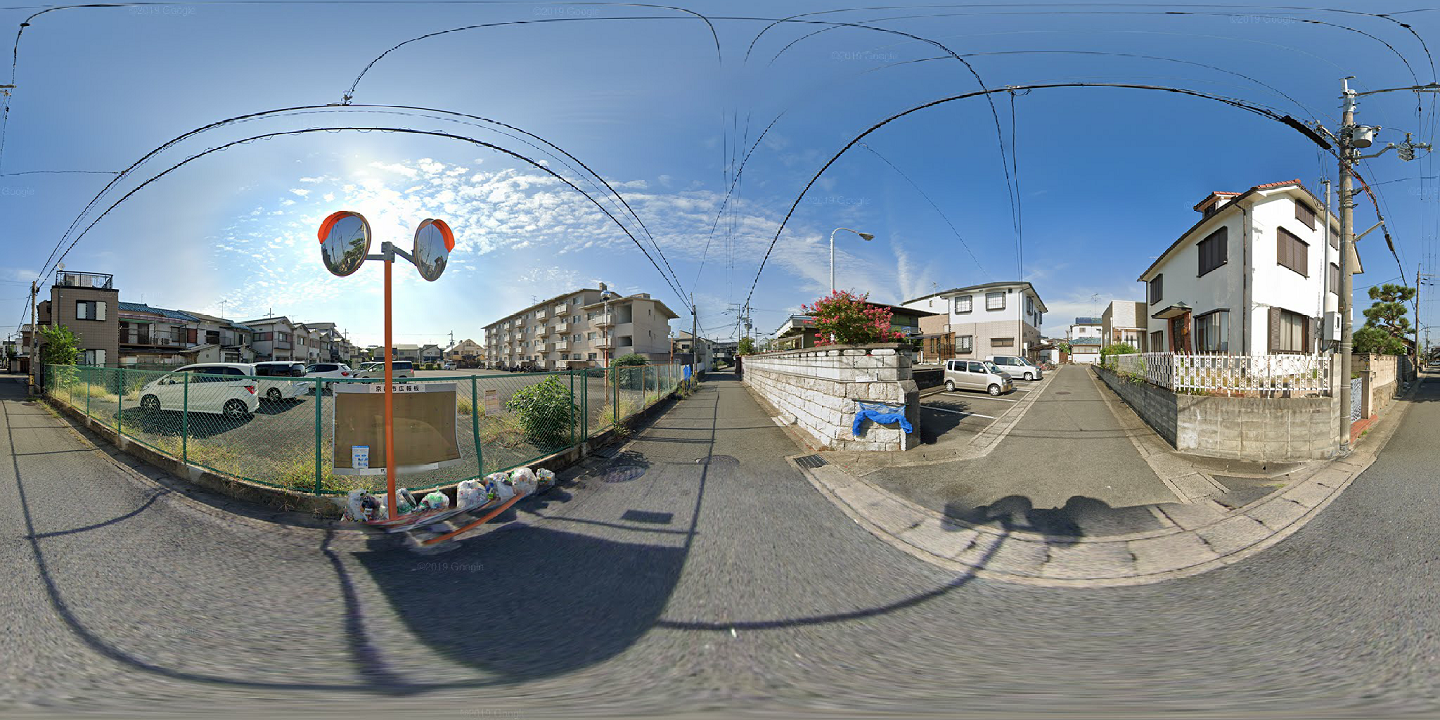}

          \vspace{2pt}
          (c) Suburb
        \end{center}
    \end{minipage}
  \begin{minipage}{0.24\hsize}
        \begin{center}
          \includegraphics[keepaspectratio,width=\textwidth]{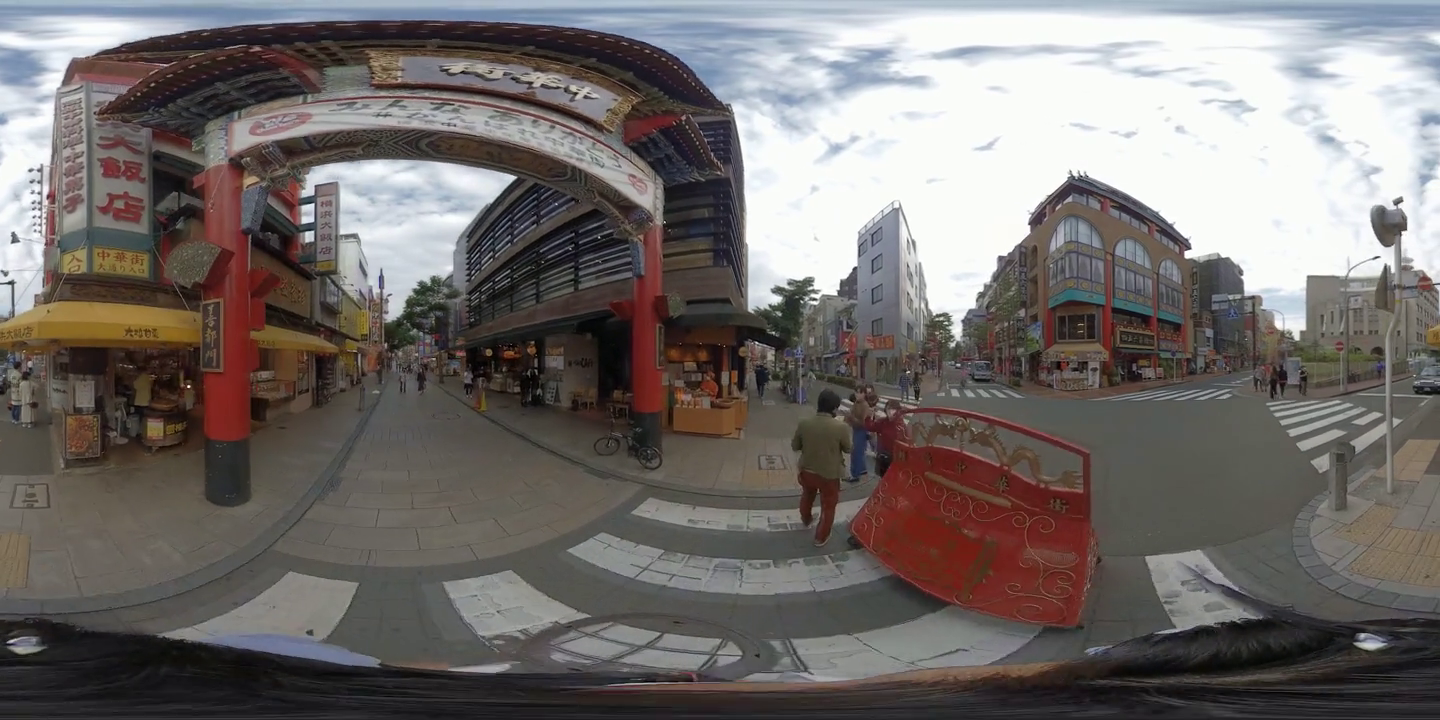}

          \vspace{2pt}
          (d) Chinatown
        \end{center}
    \end{minipage}
  \caption{Image samples from each area. Campus has large wide roads and squares and ambiguous intersections. In Downtown, most of intersections have traffic lights and pedestrian crossings. Suburb has many intersections without such elements, but the roads are clearly separated by walls or buildings. Chinatown consist of all those features where many people come and go. We used the first three areas as training data and used Chinatown data only as test data.}
  \label{dataset_samples}
  \end{center}
\end{figure*}

\noindent \textbf{Training data for PDoT detection}: First, we extracted \omni~images from the \omni~videos in Campus and Downtown areas, obtained for Movie-Map. For each area, we chose 50 videos and selected 10 frames at equal time intervals from each video; 500 frames were obtained per area. 
%For each area, we picked up ten frames at equal time intervals from each video and acquire five hundred frames from fifty videos in total. 
The two areas have different characteristics as observed in~\Fref{dataset_samples}. 
For example, on Campus, an intersection on Campus is defined ambiguously, as mentioned previously. In Downtown, streets are demarcated precisely by buildings, with maintained crosswalks or traffic lights, along with dynamic changes such as moving cars in a scene.

The maximum number of Movie-Map videos have been shot by a photographer walking on the sidewalk in a wide street. Therefore, the intersection detector should be trained in the same domain, rather than on videos shot by car-mounted cameras such as in GSV~\cite{street_view}. However, as the diversity of the intersections in the two areas mentioned previously was insufficient for generalizing the intersection identification, we supplemented it by adding another training dataset, Suburb-GSV images. The added area contains well-maintained streets but scarce traffic lights and signs; moreover, the dataset contains a few scenes without any walls or buildings and low textures. We collected 500 suburb intersection \omni~image from GSV and used them for training in addition to the images from Campus and Downtown.

\begin{figure}[t]
  \begin{center}
  \includegraphics[keepaspectratio,scale=0.3]{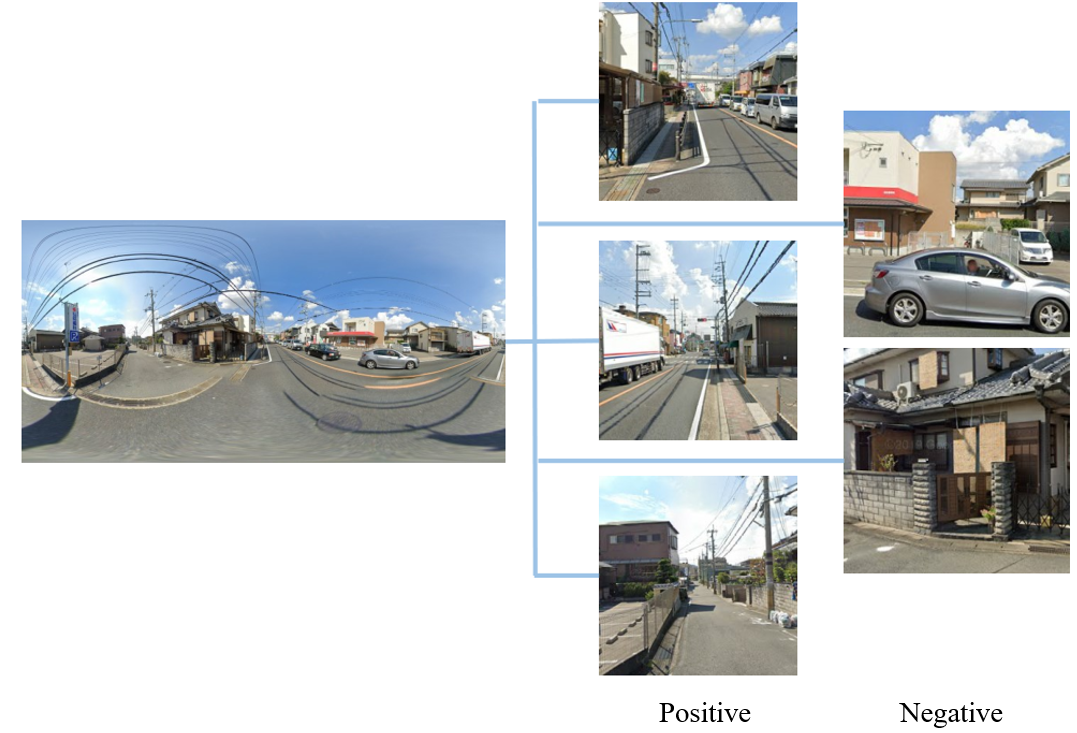}
  \caption{Illustrated strategy for generating positive and negative examples from a \omni~image for training the PDoT classifier. Positive examples are sampled as normal-field-of-view (NFoV) perspective images roughly centered at every annotated PDoT. Negative examples are sampled as NFoV perspective images centered at two adjacent PDoTs so that it contains no PDoT in its FoV.}
  \label{fvpp_dataset}
  \end{center}
\end{figure}

Next, for each key interaction frame included in this frame set, PDoTs were manually annotated on a user interface. 
Specifically, a worker was asked to indicate the PDoT in a frame with icons specifying a single or ''omnidirectional'' PDoT, where all the directions are assumed to be PDoT, for example, in a plaza on the university campus. After completing all the annotations, we cropped the NFoV (45\textdegree in horizontal and vertical views) images approximately centered at PDoT from the \omni~image as positive examples. It should be noted that, to inject noises in the training examples, we did not crop the NFoV image strictly centered at the PDoT and 
randomly shifted it up to 5 degrees. After the extraction of the positive examples from a single frame, the negative examples were similarly extracted from the same frame as NFoV images centered between two adjacent PDoTs. Notably, this sampling procedure can be applied to both key intersection and non-intersection frames (\ie, non-intersection frames have their PDoTs in the forward and backward directions). As shown in~\Fref{fvpp_dataset}, this procedure basically generates slightly more positive than negative examples because a few negative candidates are discarded when the inner angle between two adjacent PDoTs is less than 45\textdegree (\eg, Y-junction; otherwise, the negative example must contain PDoT). To equalize the number of positive and negative examples, we added additional negative examples from other random regions. Consequently, we obtained 3414 positive examples (\ie, 907 in Campus, 971 in Downtown, and 1536 in Suburb) and 3274 negative examples (859 in Campus, 882 in Downtown, and 1510 in Suburb). These samples were precisely labeled and diverse in terms of landscape.

\noindent \textbf{Training data for direct intersection classification}: 
While our proposed method identifies an intersection based on the number of PDoTs in a single  \omni~frame, the n\"aive network architecture accepts \omni~image as input and directly predicts whether the frame is an intersection. 
%To evaluate this approach, We also generated training data for this n\"aive task.
%In order to show the superiority of our method compared to the previously mentioned method that learns the relationship between a direct 360° image and whether it is an intersection or not, 
We also created training data to train the n\"aive method. Because we required a larger number of samples of images for training in the direct method, more frames were sampled from the video set. 
Rather than using Suburb-GSV for the PDoT method, we used videos shot in almost the same area. 
Thus, except Suburb and Suburb-GSV, the same sources (Campus and Downtown) were used for training the PDoT and direct methods. Both the methods used the same testing data.

%Note that we additionally took the amount of route videos in where the images were taken for Suburb area.

Specifically, equipped with the annotation of key intersections, we re-assigned the {\it soft} intersection label to every 10 frames in the videos based on the frame positions from the key intersection frame. We shot walk-around videos in Suburb similar to that in Campus and Downtown areas. Frames within $0.5$ s walking distance (\eg, $15$ frames in the $30$-fps video) from the key intersection frame were labeled as one (positive) and frames with more than two second walking distance as zero (negative); the frames decreased linearly from one to zero between $0.5$ s and two second walking distances. Because a majority of the frames were negative examples (\ie, far more than two second walking distance from the key intersection frame therefore labeled as ``False''), we balanced the portions of positive and negative examples by only extracting $p$ percentage of negative ones. We empirically found that $p=20\%$ is the optimal percentage, which means that 80 percentage of negative samples were discarded. Finally, we obtained 544 positive examples, 1225 soft-labeled examples (\ie, labeled between zero and one), and 8159 negative examples. The total distribution of (positive/soft-labeled/negative) examples for each area was (198/420/2188) for Campus, (185/457/3656) for Downtown, and (161/348/2315) for Suburb. Notable, The labels were treated as a continuous probability distribution.

\subsection{Test data generation}
The same test task was employed for both the PDoT-based and n\"aive direct approaches: intersection identification from a single 360\textdegree image. We prepared $50$ intersection and non-intersection frames, respectively, for each of the three areas and added a completely new scene called Chinatown to validate the ability of the network to generalize to unknown scenes. 
The test data resulted in a total of $400$ frames. Notably, all the test images were not included in our training data.

\section{Experimental Results}
We evaluated our PDoT-based \omni image intersection identification method (our method) on our iii360 test dataset. We compared its performance against that of the n\"aive direct method (Baseline), which we implemented as a binary classification. It should be noted that the qualitative result examples of intersection identification have been demonstrated in the supplemental paper.

\subsection{Implementation details}
Our method and Baseline were implemented using the PyTorch framework~\cite{pytorch}. The backbone architecture for both the methods was Resnet-50~\cite{resnet}, pre-trained on ImageNet~\cite{imagenet} except for the final classification layers and fine-tuned on the iii360 dataset. This implies that the only difference between these two networks was the input and output information. The input for our method was an NFoV image cropped from a target~\omni~image, whereas that for the Baseline was the~\omni~target image itself. The output was the binary labels w.r.t PDoT or intersection. Both networks accepted images resized to $224\times 224$ as input and were trained and tested on a machine with single NVIDIA TITAN Xp with 12GB of GPU memory. 
For optimization, we used Adam~\cite{adam} with a learning rate of 0.001 and batch size of $8$. The number of epochs was set to $300$ for all network and training set combinations. For data augmentation, horizontal flipping, color jitter, and random erasing were applied.

To identify the intersection, our method merges the multiple PDoT predictions as follows. First, in a provided target~\omni~image, horizontally non-overlapping eight perspective images of $45^\circ\times 45^\circ$ FoV are cropped (in a random start direction). Then, each NFoV image is individually fed to the PDoT prediction network, and the target~\omni~image is identified as an intersection if when more than three PDoT predictions are positive.

\subsection{The effect of hyperparameters}
We investigated the effect of the number of perspective images cropped from 360° images. We varied it within 8, 16 and 32 (\ie, 45\textdegree~, 22.5\textdegree~and 11.25\textdegree~FoV, respectively) and compared the prediction performance by training the network on Campus and testing on Chinatown datasets. The result is shown in~\Tref{view_score}, and we found the result of 8 views is the best.
%There wasn't much difference, and we decided to use 8 views.

In addition, in order to evaluate the effect of the resolution of input images in the direct method, we compared the prediction accuracy by training the network on Campus and testing on Chinatown datasets by varying the image size among $224\times224$, $448\times448$ and $896\times896$. ~\Tref{resolution_score} demonstrated that the difference of input image size is not influential to the performance.

\subsection{Quantitative results}

%\begin{figure*}[t]
%  \begin{center}
%  \includegraphics[keepaspectratio,width=\textwidth]{figure/result_examples5.png}
%  \caption{Examples of intersection identification results for different areas. "True" means that the image was classified an intersection, "False" means that it was classified as non-intersection. The first row shows the result about the intersection images and the second row shows the result about non-intersection images. The third row shows examples that the proposed method failed to assign correct intersection labels. Please see detailed analysis in Sec. 5.2.}
%  \label{result_examples}
%  \end{center}
%\end{figure*}

\begin{table}[t]
  \label{fvpp_score}
  \centering
  \begin{tabular}{ccccc:c}
    \hline
      Method & Dataset & \multicolumn{4}{c}{Test set domain}  \\
      & & Campus & Downtown & Suburb & Chinatown \\
      \multirow{4}{*}{PDoT} & Campus & 0.78 & 0.82 & 0.70 & 0.65 \\
      & Downtown & 0.67 & 0.88 & 0.75 & 0.87 \\
      & Suburb-GSV & 0.68 & 0.78 & 0.81 & 0.57 \\
      & Three datasets & 0.74 & 0.86 & 0.81 & 0.88 \\ \hline
      \multirow{4}{*}{Direct} & Campus & 0.65 & 0.54 & 0.51 & 0.57 \\
      & Downtown & 0.52 & 0.81 & 0.68 & 0.67 \\
      & Suburb& 0.49 & 0.72 & 0.78 & 0.66 \\
      & Three datasets & 0.57 & 0.68 & 0.69 & 0.65 \\  \hline
  \end{tabular}
  \caption{Intersection prediction results.}
\end{table}

\begin{table}[t]
  \begin{minipage}{0.48\hsize}
  \centering
  \begin{tabular}{cc}
    \hline
      Views & Accuracy \\ \hline
      8 & 0.65 \\
      16 & 0.58 \\
      32 & 0.55 \\ \hline
  \end{tabular}
  \caption{Result of proposed method with different views of cropped images. Training data was the Campus area data and Test data was the Chinatown area data in iii360.}
  \label{view_score}
  \end{minipage}
  \hfill
  \begin{minipage}{0.48\hsize}
  \centering
  \begin{tabular}{cc}
    \hline
      Resolution & Accuracy \\ \hline
      $224\times 224$ & 0.57 \\
      $448\times 448$ & 0.55 \\
      $896\times 896$ & 0.58 \\ \hline
  \end{tabular}
  \caption{Result of Direct method with different resolutions. Training data was the Campus area data and Test data was the Chinatown area data in iii360.}
  \label{resolution_score}
  \end{minipage}
\end{table}

We compared our method with Baseline having different combinations of training/test data. In addition to training the networks on examples from individual areas, we attempted training on all the examples from all areas. Notably, we did not evaluate PDoT prediction accuracy because this was not our main task; instead, we assigned more importance to the intersection identification performance.

The comparative intersection identification accuracy results are shown in~\Tref{fvpp_score}. It is evident that the prediction accuracy of our method is better than that of Baseline for all the combinations of training and test areas, indicating that our PDoT-based algorithm consistently outperforms the n\"aive direct approach of the Baseline algorithm.

As expected, the performance was better when the training and test examples were drawn from the same area, compared to when drawn from different areas. Interestingly, the network trained on Suburb-GSV still performed satisfactorily on the test Suburb dataset despite the completely different device setups for obtaining both types of data, indicating that a difference in the image acquisition setup exerts a lesser effect than that in the area where the image was captured. The $0.87$ prediction accuracy demonstrated by the network trained on Downtown and tested on Chinatown also supports this observation as both these areas have several common characteristics, such as movement of people through narrow and intricate shopping streets. Conversely, the network trained on Campus did not perform well on Chinatown where both areas have less shared characteristics. Although the excellent performance of the network on Downtown even after training on Campus appears counter-intuitive, the average prediction accuracy on Downtown is consistently high and simply indicates that intersection identification in this area is relatively easier than in other areas with more diversity in the appearance of intersections. It is not surprising that the network trained on all the training examples performed slightly worse than when the training and test examples were drawn from the same area. Instead, this result indicates that the network is capable of covering a variety of areas by drawing training images of intersections from them.

\section{Conclusion}

In this paper, we presented a new algorithm to identify an intersection from a single~\omni \, image. We propose a PDoT-based method that identifies intersections by the number of possible directions of travel, rather than directly identifying the \omni~image with a binary classifier. For training our PDoT detection network and evaluating the method, we constructed a new large-scale \omni~Image Intersection Identification (iii360) dataset. Although the original motivation of our work was the automatic intersection segmentation of Movie-Map videos, we believe our method is also applicable in other fields, such as autonomous driving and robot navigation.

As our network architecture was quite basic (ResNet-50) and the training examples were not that large in number, improving the network architecture and increasing the training dataset size is an important future direction. Another important future work is to demonstrate the effectiveness of our method in actual Movie-Map application.

\section{Acknowledgement}

This work is partially supported by JSPS KAKENHI 21H03460, JST-Mirai Program JPMJMI21H1 and VTEC Lab.

\newpage

\bibliography{all}
\end{document}